\documentclass{article}
\usepackage{spconf,amsmath,graphicx}
\usepackage{amssymb,amsmath,epsfig}
\usepackage{lipsum,color,multirow,url}
\usepackage{arydshln, algorithm, algpseudocode}
\usepackage{paralist}

\title{An Empirical Evaluation of Zero Resource Acoustic Unit Discovery}
\makeatletter
\def\name#1{\gdef\@name{#1\\}}
\makeatother
\name{ \em Chunxi Liu \sthanks{Both authors contributed equally to this work.}$^{\dagger}$,  Jinyi Yang\footnotemark[1]$^{\ddagger}$, Ming Sun$^{\star}$, Santosh Kesiraju$^{\|}$,  Alena Rott$^{\P}$,   {Lucas Ondel}$^{\S}$, Pegah Ghahremani$^{\dagger}$,  \\  \em Najim Dehak$^{\dagger}$, {Luka{\v{s}} Burget}$^{\S}$ , Sanjeev Khudanpur$^{\dagger}$ } 

\address{
$^\dagger$Johns Hopkins University  \ \   $^\ddagger$University of Chinese Academy of Sciences  \  \  $^\star$Amazon \\
  $^\|$International Institute of Information Technology       $^\P$Stanford University        $^\S$Brno University of Technology
}
\begin{document}
\ninept
\maketitle
\begin{abstract}
Acoustic unit discovery (AUD) is a process of automatically identifying a categorical acoustic unit inventory from speech and producing corresponding acoustic unit tokenizations. AUD provides an important avenue for unsupervised acoustic model training in a zero resource setting where expert-provided linguistic knowledge and transcribed speech are unavailable. 
Therefore, to further facilitate zero-resource AUD process, in this paper, we demonstrate acoustic feature representations can be significantly improved by (i) performing linear discriminant analysis (LDA) in an unsupervised self-trained fashion, and (ii) leveraging resources of other languages through building a multilingual bottleneck (BN) feature extractor to give effective cross-lingual generalization. 
Moreover, we perform comprehensive evaluations of AUD efficacy on multiple downstream speech applications, and their correlated performance suggests that AUD evaluations are feasible using different alternative language resources when only a subset of these evaluation resources can be available in typical zero resource applications.

\end{abstract}
 \section{Introduction}
\label{sec:intro}
Standard supervised training of automatic speech recognition (ASR) systems typically relies on transcribed speech audio and pronunciation dictionaries. However, for a large majority of the world's languages, it is often difficult or even almost impossible to collect enough language resources to develop ASR systems with current standard ASR technology~\cite{Besacier2014automatic}. 
Therefore, developing speech technologies for a target language with zero expert-provided resources in that language becomes a significant challenge. 

Recent zero resource efforts focused on phonetic discovery, or acoustic unit discovery (AUD), have made important progress in fully unsupervised acoustic model training and performing subword unit tokenization~\cite{lee2012nonparametric, ondel2016variational}. 
In~\cite{lee2012nonparametric}, a Dirichlet process hidden Markov model (DPHMM) framework is formulated to simultaneously perform three sub-tasks of segmentation, nonparametric clustering and sub-word modeling, and the spoken term detection task is used to evaluate the learned sub-word models. 
\cite{ondel2016variational} also presents a nonparametric Bayesian framework to solve the same problem of unsupervised acoustic modeling with three major differences: (i) the Gibbs Sampling (GS) training algorithm is replaced with Variational Bayesian (VB) inference, which allows parallelized training amenable to large scale applications, (ii) a phone-loop model with a mixture of HMMs (each phone-like acoustic unit is modeled by a HMM) is seen as a single HMM and thus does not require sub-word boundary variables, and (iii) normalized mutual information (NMI) between the hypothesized acoustic unit sequences and orthographic phoneme transcripts is used to evaluate the modeling efficacy. 

As being unknown to the guidance of word transcripts in zero-resource scenarios, effective acoustic front-end processing becomes particularly critical to uncover the phonetic salience by the acoustics themselves. In supervised ASR system, linear discriminant analysis (LDA)~\cite{fisher1936use} is often employed to exploit substantial contextual information, and the target class labels for LDA can be context-dependent triphone states
given by the forced alignments of speech transcripts that are unavailable in zero resource setting. In this paper, we explore applying similar LDA strategy but with target labels acquired by the first-pass acoustic unit tokenizations, which is considered as a self-supervised fashion to solve the unknown label problem. Previous work in \cite{heck2016unsupervised} also exploits such unsupervised LDA to support Dirichlet process Gaussian mixture model (DPGMM) based clustering although being limited to frame-level clustering without acoustic unit-level segmentation. 

To date language-independent bottleneck (BN) features have been demonstrated as effective speech representations in improving ASR accuracies~\cite{yu2011improved, vesely2012language}. In our study, we explore a state-of-the-art multilingual time delay neural network (TDNN) technique to generate robust cross-lingual acoustic features in zero-resource setting, and the hope is that as one moves to new languages, this data driven feature extraction approach will work as-is, without having to redesign feature extraction algorithms.

In this paper, we employ the AUD framework in~\cite{ondel2016variational} and investigate the efficacy of incorporating LDA and multilingual BN TDNN techniques to AUD. 
Given the two distinct evaluations in~\cite{lee2012nonparametric, ondel2016variational}, we proceed by conducting not only an intrinsic measure of assessing the NMI between model hypothesis and true reference, but also an extrinsic measure of AUD's utility to downstream speech tasks. 

Past studies in \cite{zhang2009unsupervised, chen2015parallel} demonstrated the effectiveness of posterior features based on automatically derived acoustic structures by spoken term detection and phoneme discrimination tasks, while being limited to frame-level clustering and loss of phonetic temporal information. In contrast,~\cite{lee2012nonparametric} succeeded in computing posteriorgram representations over the learned sub-word units, capturing the phonetic context knowledge. In this paper, we also exploit the feature representation of posteriorgrams across acoustic units learned from our AUD procedure, and test by a unified evaluation framework proposed in~\cite{michael2010rapid, jansen2013summary} that quantifies how well speech representations enable discrimination between word example pairs of the same or different type, which is referred to as the same-different task and characterized by average precision (AP).
\cite{michael2010rapid} demonstrates almost perfect correlation between such AP and phone recognition accuracies of supervised acoustic models; therefore, we would like to investigate if such AP 
can also be a proxy for the unsupervised AUD accuracies, such that we can still evaluate AUD efficacy in the zero-resource condition that no orthographic phoneme transcripts for NMI measure are available but only word pairs. 
Since such word pairs can not only be obtained from manual transcripts, also from unsupervised spoken term discovery systems without relying on any language-specific resources~\cite{jansen2011efficient, lyzinski2015evaluation}, at little cost compared with expensive phoneme transcripts. 

Finally, previous work like~\cite{siu2014unsupervised, dredze2010nlp} presented the success of using unit- or word-level acoustic patterns discovered from fully unsupervised setting to provide competitive performance in spoken document topical classification and clustering, we also explicitly measure our AUD utility of learning document representations in this study.   

\section{Improving Feature Representation Learning for acoustic unit discovery}
\label{sec:aud}

\begin{figure}[t]
\begin{minipage}[b]{1.0\linewidth}
  \centering
  \centerline{\includegraphics[width=5.0cm, scale=0.33]{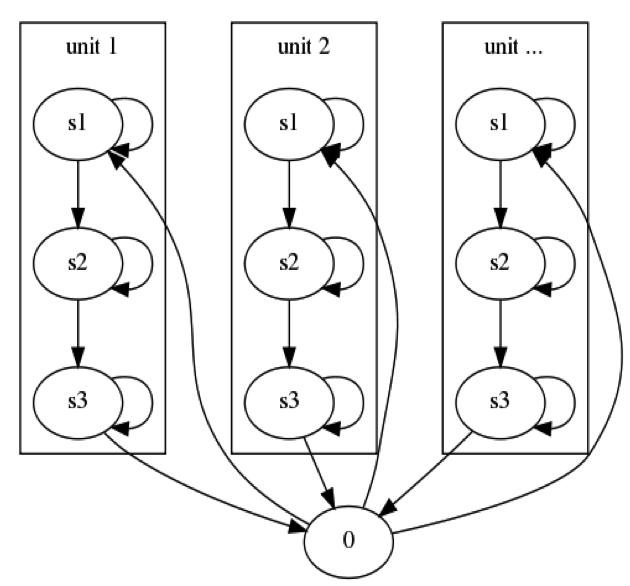}}
\end{minipage}
\vspace{-0.70cm}
\caption{AUD phone-loop model with an infinite number of units and each unit modeled by a Bayesian GMM-HMM.}
\vspace{-0.52cm}
\label{fig:phoneloop}
\end{figure}

AUD is to discover repeated acoustic patterns in the raw acoustic stream and learn speaker independent acoustic models for each unique acoustic unit. We employ the same nonparametric Bayesian framework as~\cite{ondel2016variational}. A phone-loop model is developed as shown in Figure~\ref{fig:phoneloop}, and each unit is modeled as a Bayesian GMM-HMM. Under Dirichlet process framework, we consider the phone-loop as an infinite mixture of GMM-HMMs, and the mixture weights are based on the stick-breaking construction of Dirichlet process. Following~\cite{blei2006variational}, the infinite number of units in the mixture is approximated by a truncation number $T$,  giving zero weight to any unit greater than $T$. 

Model parameters are fully Bayesian, with corresponding prior and posterior of conjugate distributions for each parameter. 
The Variational Bayesian (VB) inference (seen as an extension of the expectation-maximization algorithm that computes posterior distributions of both model parameters and latent variables)~\cite{jordan1999introduction} is used to train the full Bayesian models. 
We initialize the hyperparameters for the prior distributions,  and the posterior distributions are initialized the same as their prior distributions before the first training iteration starts. During each iteration, sufficient statistics are computed and accumulated to update the model posterior distributions. 
We can treat such mixture of GMM-HMMs as a single unified HMM with loop transitions, and thus the segmentation of the data can be performed using standard forward-backward algorithm in an unsupervised fashion. Parallelized training is conducted and convergence monitored by computing a lower bound on the data log-likelihood. 
After VB training, during evaluation we use Viterbi decoding algorithm to obtain acoustic unit tokenizations of the data, or forward-backward algorithm to produce posteriorgrams across the learned acoustic units.

\subsection{LDA with Unsupervised Learning}


We first parameterize the acoustic data into Mel-frequency cepstral coefficients (MFCCs) or BN features and apply Cepstral mean and variance normalization (CMVN), perform a first-pass AUD training over such raw acoustic features,
obtain acoustic unit HMM state tokenizations of the data, i.e., 1-best HMM state-level decode for each acoustic frame, and use the resulting state-level labels as the class labels for LDA. 

To apply LDA, additional context frames after CMVN are stacked to around the center frame. LDA is then performed on this higher-dimensional, context-rich representation. We apply the resulting LDA transformation to project the context-rich raw acoustic features back into a lower dimensional representation. These vectors after CMVN are subsequently used for a second-pass AUD training. Note that, for the second-pass VB training on the LDA-based features, rather than starting from scratch, we can first use the models learned from first-pass training to compute certain sufficient statistics that can be transferred regardless of different front-end features, and use them to update the model posteriors just for the first iteration; e.g., we can transfer the MFCC/BN-based statistics of accumulated posteriors of certain latent variables (acoustic unit, HMM state or GMM component), and use them to re-estimate the posterior's parameters in the first iteration of LDA-based training, by assuming certain acoustic structures discovered by the first-pass model being more accurate than those by our prior models.

\subsection{Cross-lingual Generalization of Multilingual BN Network}

In our multilingual BN training recipe, we use the TDNN architecture with parallel GPU training (using up to 8 GPUs) as described in~\cite{peddinti2015time} with two major extensions. First, hidden layers with ReLU nonlinearity are shared across languages (ReLu dimension 600, i.e., the output dimensions of the weight matrices), while separate language-specific final output layers with context-dependent triphone state targets are used for each different language. 
Second, an additional 42-dimensional bottleneck layer is added just before the final output layers, giving 6 hidden layers in total. 
Moreover, 3-fold training data augmentation with speed perturbations of 0.9, 1.0 and 1.1 are used.
Each mini-batch of training data is randomly sampled based on the relative amounts of acoustic data in different languages, and any data of each language is used only once in one epoch. 
40-dimensional MFCCs (without cepstral truncation~\cite{peddinti2015time}) augmented with 3-dimensional pitch and probability of voicing features are used as inputs to the network. 

We developed our TDNN-based BN training and validation using multiple languages in both hybrid and tandem HMM-based ASR systems, while being unknown to the target language on which we perform AUD. We assume the word error rate reductions in our BN-based ASR tasks will translate into more effective cross-lingual generalization of our BN techniques on unseen target language, in turn, facilitating more accurate AUD.

%
%
\vspace{-0.1cm}
%
%
%
\section{Evaluating Acoustic Unit Discovery}
\label{sec:evaluate}
\subsection{NMI against Orthographic Phoneme Transcripts}

After VB training of the Bayesian AUD models, to evaluate the quality of the automatically learned acoustic models, we first obtain acoustic unit tokenizations, i.e., 1-best HMM unit-level decode, of the development data on which AUD training is performed; alternatively, we can also use the learned models to obtain tokenizations of any evaluation data that the models do not see during training. Then we align the decoded acoustic unit sequence $\bf{Y}$$= Y_1, ..., Y_N$ with reference phoneme sequence $\bf{X}$$= X_1, ..., X_M$, and each $Y_j (1 \leq j \leq N$) is aligned to a $X_i (1 \leq i \leq M$), based on which the mutual information $I(\bf{X}; \bf{Y})$ is computed. We normalize it by the entropy $H(\bf{X})$ of $\bf{X}$, giving the normalized mutual information $NMI = I(\bf{X}; \bf{Y}) / H(\bf{X})$. $NMI=0$ means $\bf{Y}$ carries no information about $\bf{X}$, and $NMI = 1$ means $\bf{Y}$ perfectly predicts $\bf{X}$. 

\subsection{Same-Different Evaluation}
\label{sec:typestyle}

To evaluate AUD models, we can also apply them to a data set by using forward-backward algorithm to compute posterior distributions across the learned acoustic units over time. Thus, any word segments required by the same-different task can be given such HMM unit-level or state-level posteriorgram features. For each word pair, we compute a pairwise normalized dynamic time warping (DTW) distance with symmetric KL-Divergence as frame-level distance metric; since our acoustic features are posterior distributions, symmetric KL divergence are demonstrated superior to cosine distance for posterior features~\cite{michael2010rapid}. The pairwise normalized DTW distance is further used as a same/different classifier score; if the score is lower than some threshold $\tau$, we declare this word pair corresponds to the same word type. As we sweep the threshold $\tau$, we can obtain a standard precision-recall curve, under which the area is computed as the average precision (AP). 
In such means, we investigate if the better posterior estimates across automatically derived acoustic categories in AUD procedure can translate into the improved discriminability of separating same word type pairs from different word type pairs.

\subsection{Spoken Document Classification and Clustering}

Spoken document topical classification/identification (ID) is to classify a given document into one of the predefined set of topics or classes. Typically, documents are characterized based on a bag-of-words multinomial representation~\cite{Hazen:2011:MCE_topic_ID}, or a more compact vector given by probabilistic topic models~\cite{Blei:2003:LDA}. 
To evaluate the quality of the acoustic unit tokenizations of spoken documents, we employ the document representations as bags of acoustic units. For such classification task with topic labeled training data, we use stochastic gradient descent based linear SVM~\cite{Shalev:2007:SVM, scikit-learn} as our multi-class classifier training algorithm, with hinge loss and $\mathcal{L}^1$ norm regularization. 

In the case that no topic labels are available, we can still perform unsupervised document clustering by the bags of acoustic units representation. We would like to investigate if reasonable clustering performance can be obtained without using manual or automatic transcript from supervised acoustic models but only unsupervised AUD.
Following~\cite{dredze2010nlp}, we use the clustering algorithm of globally optimal repeated bisection~\cite{karypis2010cluto}.


%
%
\section{Experiments}
\label{sec:exp}
\subsection{Experimental Setup}

For our experiments we use the Switchboard Telephone Speech Corpus~\cite{godfrey2010switchboard}, a collection of two-sided telephone conversations with a single participant per side. 
Following the data set split strategy in~\cite{dredze2010nlp}, we use the same development and evaluation data set as~\cite{dredze2010nlp}. There are 360 conversation sides of six different topics (recycling, capital punishment, drug testing, family finance, job benefits, car buying) in the development data set of 35.7 hours of audio. Each conversation side (seen as a single document) has one single topic, and each topic has equal number of 60 sides of conversations. Similarly, there are another different six topics (family life, news media, public education, exercise/fitness, pets, taxes) evenly across the 600 conversation sides of evaluation data set (61.6 hours of audio). Unsupervised VB training of acoustic unit models are performed on the development set (10 iterations); after the unsupervised learning, we apply the learned acoustic unit models to obtain the acoustic unit tokenizations of both development and evaluation sets.

For AUD model definitions, we use the truncation $T=200$, which implies maximum 200 different acoustic units can be learned from the corpus. For each acoustic unit, we use a HMM of 3 emission states with a left-to-right topology and 2 Gaussians per state. Other hyperparameter values are the same as~\cite{ondel2016variational}. 

To compute NMI, we first use a supervised ASR system trained on Switchboard training corpus (about 300 hrs) to obtain forced aligned phoneme transcripts as our reference transcripts. During scoring, we define the distance between an output acoustic unit token and a reference phoneme token as the time frame difference between the center frames of two tokens; in doing so, each acoustic unit token is assigned to a closest reference phoneme token based on the distance metric defined.
As shown in Table~\ref{tab:results}, the number of units in the tokenizations of a dataset is determined as the number of unique units that occur in any of the 1-best Viterbi decode of that dataset; thus, truncation $T=200$ is the ceiling number, and it is possible that unit numbers differ between development and evaluation data since all 200 unit models are used during decoding process.

\setlength{\tabcolsep}{0.24cm}
\renewcommand{\arraystretch}{1.10}
\begin{table*}[t]
\vspace{-0.5cm}
\caption{\label{tab:results} {\it  AUD Performance evaluated by NMI, same-different task, document classification and clustering on Switchboard}}
\vspace{1.4mm}
\centerline{ 
\begin{tabular}{c|c|c|c|c|c|c c }
\hline \hline
 &  Acoustic Features   &   &     & Average   &      Document Classification   &  \multicolumn{2}{ |c  }{ Document Clustering     }   \\
       Dataset      &            AUD is based on               &         \# units         &        \% NMI         &     Precision    &       Accuracy          &      Purity   &          B-Cubed F1       \\
\hline \hline
\multirow{1}{*}{ }   &   MFCC            &  145       & 21.59  & 0.247   & 0.3083 $\pm$ 0.0908   &  0.2268  $\pm$  0.0015   & 0.1817  $\pm$  0.0008 \\    
	Development	&        MFCC w/ LDA    &   145       & 24.55  &    0.251      &   0.4361   $\pm$ 0.0692     &    0.2354  $\pm$ 0.0026    &    0.1855    $\pm$  0.0006 \\   \cdashline{2-8}[2.0pt/0.5pt]
\multirow{1}{*}{ Data}      & BN                   &    184     &      28.20      &  0.343    &     0.7028   $\pm$    0.0796      &     0.2446     $\pm$  0.0018   &   0.1949  $\pm$ 0.0008 \\   
                                            &   BN w/ LDA   &     184      &   29.13       &      0.359     &     0.7167  $\pm$   0.0733      &    0.2553  $\pm$  0.0102       &   0.2023  $\pm$  0.0047        \\ 
\hline \hline
\multirow{1}{*}{ }    &  MFCC              &   144  &   21.20        &       0.224   &      0.4633  $\pm$  0.0702      &    0.2388     $\pm$   0.0010    &   0.1899    $\pm$     0.0001   \\  
		Evaluation		& MFCC w/ LDA   &    144  &      24.07  &   0.219   &  0.4833   $\pm$   0.0477   &     0.2426   $\pm$  0.0031   &        0.1893         $\pm$   0.0005  \\    \cdashline{2-8}[2.0pt/0.5pt]
                         Data   & BN                  &     184      &    28.01     &     0.303     &         0.7167     $\pm$   0.0350     &      0.2398  $\pm$ 0.0069        &   0.1983 $\pm$  0.0032    \\ 
                                                   & BN w/ LDA       &       184        &     28.84      &    0.329  &    0.7300  $\pm$ 0.0567    &        0.2373   $\pm$  0.0037       &   0.2140  $\pm$  0.0035  \\
\hline \hline
\end{tabular}}
\vspace{-0.5cm}
\end{table*}

\subsection{Feature Extraction Using LDA and Multilingual BN}


For AUD experiments, we use manual segmentations provided by the Switchboard corpus to produce utterances with speech activity, and speech utterances are further parameterized either as 39-dimensional MFCCs with first and second order derivatives, or 42-dimensional BN features, with CMVN applied per conversation side. 

11-frame context windows of raw acoustic features (MFCCs or BN features) with CMVN are stacked to represent the center frame (equal left and right context frames as 5), and used as the LDA inputs. 
Using truncation parameter $T = 200$ and 3 HMM emission states yields 600 possible unique HMM state labels for the first-pass tokenization of development data. These state labels are used as LDA class labels.  
We accumulate LDA statistics and estimate the transformation matrix from development data, and apply the resulting LDA transformation to both development and evaluation data, reducing the spliced raw acoustic features into 40 dimensions for each frame. Then we proceed with second-pass AUD training based on the 40-dimensional LDA features. We reuse the sufficient statistics of certain latent variables (i.e., accumulated posteriors of each acoustic unit, HMM state transitions and GMM component) that are computed by the first-pass AUD model on raw features, for updating the model posterior distributions in the first iteration of second pass training; we find empirically, this procedure outperforms conducting the second-pass training on LDA features from scratch (i.e., initializing posterior distributions the same as their priors). 

Using the Kaldi toolkit~\cite{povey2011kaldi}, we conduct our multilingual TDNN-based BN training with 10 language collections provided in the IARPA Babel Program (IARPA-BAA-11-02): Assamese, Bengali, Cantonese, Haitian, Lao, Pashto, Tamil, Tagalog, Vietnamese and Zulu. 10-hour transcribed speech of each language is used for training. We first evaluated our multilingual BN recipe for ASR experiments using this Babel corpus, and observed modest WER improvements in the hybrid multilingual TDNN system by using other languages to supplement the training data of test language, and more robust WER improvements in the tandem TDNN system with spliced multilingual TDNN-based BN features and MFCCs. Detailed discussion of ASR results is beyond the scope of this paper. Particularly, we are interested in learning speech representations with effective cross-lingual generalization to an unseen language as in a zero-resource setting where AUD is typically performed. 
The multilingual TDNN-based BN training recipes will be available in the Kaldi code repository~\cite{povey2011kaldi} as an open-source capability of language-independent BN feature extraction. 

\subsection{AUD Evaluations}

For same-different tasks, from the time aligned word transcriptions, we extracted all word examples that are at least 0.50 s in duration and at least 6 characters as text from development and evaluation data set respectively. Development set produces approximately 11k word tokens, 60.8M word pairs of which 96.8k have the same word type. Evaluation data has 19k tokens and 186.9M word pairs with 281.8k pairs having the same word type. 200  ($T = 200$) dimensional AUD posteriorgram features across HMM units are produced as inputs to DTW scoring function.  

For document classification, we use the acoustic unit trigram representation,
and we scale each trigram feature value by the inverse document frequency, referred to as TFIDF features. We further normalize each feature vector to $\mathcal{L}^2$ norm unit length. To be comparable with experimental results in~\cite{dredze2010nlp}, classification accuracies are reported based on 10-fold cross validation, and average performance with standard deviations reported in Table~\ref{tab:results}. 

For document clustering, we use the TFIDF features of the spliced acoustic unit unigrams, bigrams and trigrams. Purity and B-Cubed F1 score~\cite{bagga1998entity, amigo2009comparison} are used as evaluation metrics. We run all clustering experiments using Cluto clustering library~\cite{karypis2010cluto}, and for each one, we use 10 different initializations and report average performance and standard deviations.

As shown in Table~\ref{tab:results}, for NMI, same-different task and document classification, both LDA and BN features produce substantial and correlated improvements, except that performing LDA on MFCCs does not seem to improve the same-different AP. Specifically, the best performance across all measures by combining LDA and BN demonstrates the complementarity between these two approaches. 
Given all the same AUD model configurations, we find the improved same-different AP or document classification accuracy often indicates the NMI improvement, which implies in a zero-resource setting, AUD evaluation can fall back to other resources if necessary, e.g., word pairs or topic labels,
which might be easier to be available or to obtain than the expensive orthographic phoneme transcripts. 

Also, as we can see, NMI only drops slightly between development and evaluation data, which shows the learned acoustic unit models can generalize well on unseen data. 

Moreover, we find directly using the raw MFCCs after CMVN as acoustic features for all word segments gives AP 0.208 on the same-different task of development data. Therefore, the significantly higher AP 0.247 provided by our AUD posterior features across acoustic unit HMMs (learned from MFCCs) demonstrates AUD posteriorgrams as effective acoustic representations.
We also find the 600-dimensional AUD posterior features across each acoustic unit HMM state can provide even higher AP as 0.267, and we leave all the applications of HMM-state based posterior features for the future work, since HMM state-level posteriors (with 3 times larger dimensions than HMM unit-level posteriors if we use 3 state HMM) have much larger computational overhead in DTW scoring.

Also, our legitimate zero-resource document classification effort with LDA- and BN-based AUD yields accuracy 0.73, which demonstrates AUD tokenizations to be effective document representations for discriminative tasks. For topic clustering in development data, there are consistently marginal gains as other measures improved. However, this trend does not well hold in evaluation data. Moreover, on the same development data we use,~\cite{dredze2010nlp} shows phone trigram features by a high-resource supervised phoneme recognizer give classification accuracy up to 0.9138, clustering purity $0.6194$ and $B^3$ F1 score $0.5256$, which indicates document processing with unsupervised phonetic information remaining a challenging task.
%
\section{CONCLUSIONS}

We present an effective AUD framework that
can be successfully improved by integrating a self-supervised LDA technique and a complementary language-independent TDNN-based BN feature extraction recipe. We demonstrate the effectiveness of AUD-based discriminative features as acoustic representations given by AUD posteriors across automatically discovered units, and as document representations given by AUD tokenizations. Moreover, we find the gains in the intrinsic NMI metric for AUD algorithm development can often be predicted by the improved efficacy of applying AUD to real speech applications like same-different task and document classification.
This suggests that in real zero-resource scenarios, as we optimize the core AUD technology, alternative evaluations by various different resources can be considered, 
which serve as zero resource efforts towards ASR technology without relying on any expert-provided linguistic knowledge. 

\vspace{-0.16cm}
\section{Acknowledgements}

Most of the work presented here was carried out during the 2016 Jelinek Memorial Summer Workshop on Speech and Language Technologies, which was supported by Johns Hopkins University via DARPA LORELEI Contract No HR0011-15-2-0027, and gifts from Microsoft, Amazon, Google, and Facebook.

\vfill\pagebreak


\bibliographystyle{IEEEbib}
\bibliography{refs}

\end{document}